\documentclass[10pt,journal,letterpaper,compsoc]{IEEEtran}

\usepackage{amsmath}
\usepackage{amssymb}
\usepackage{fancybox}
\usepackage[lofdepth,lotdepth]{subfig}
\usepackage{longtable}
\usepackage[ruled,noline,linesnumbered]{algorithm2e}
\usepackage{graphicx,wrapfig}
\usepackage{subfig}
\newenvironment{Table}
  {\par\bigskip\noindent\minipage{\columnwidth}\centering}
  {\endminipage\par\bigskip}

\newcommand{\bm}[1]{{\boldsymbol{\mathbf{#1}}}}
\newcommand{\mbf}{\bm}

\newcommand{\beq}{\begin{equation}}
\newcommand{\eeq}{\end{equation}}

\newcommand{\qq}[1]{``#1''}
\newcommand{\bieee}[1]{\vspace{-0.02in} \begin{IEEEeqnarray}{#1}}
\newcommand{\eieee}{\end{IEEEeqnarray}}
\newcommand{\la}{\leftarrow}

\newcommand{\qed}{\nobreak \ifvmode \relax \else
      \ifdim\lastskip<1.5em \hskip-\lastskip
      \hskip1.5em plus0em minus0.5em \fi \nobreak
      \vrule height0.75em width0.5em depth0.25em\fi}

\SetKwFunction{Kron}{kron\_mvprod}

\begin{document}
%
\title{Scaling Multidimensional Inference for Structured Gaussian Processes}

%
%
%
%

\author{Elad~Gilboa,
        Yunus~Saat\c{c}i,
       and~John~P.~Cunningham

\IEEEcompsocitemizethanks{
\IEEEcompsocthanksitem E. Gilboa is with the Preston M. Green Department of Electrical and System Engineering, Washington University in St. Louis. E-mail: gilboae@ese.wustl.edu
\IEEEcompsocthanksitem Y. Saat\c{c}i is with the Department of Engineering University of Cambridge, UK. E-mail: yunus.saatci@gmail.com
\IEEEcompsocthanksitem J. Cunningham is with the Department of Engineering University of Cambridge, UK. E-mail: jpc74@cam.ac.uk.}
\thanks{}
}


\IEEEcompsoctitleabstractindextext{%
\begin{abstract}
Exact Gaussian Process (GP) regression has $\mathcal{O}(N^{3})$ runtime for data size $N$, making it intractable for large $N$. Many algorithms for improving GP scaling approximate the covariance with lower rank matrices. Other work has exploited structure inherent in particular covariance functions, including GPs with implied Markov structure, and equispaced inputs (both enable $\mathcal{O}(N)$ runtime). However, these GP advances have not been extended to the multidimensional input setting, despite the preponderance of multidimensional applications. This paper introduces and tests novel extensions of structured GPs to multidimensional inputs. We present new methods for additive GPs, showing a novel connection between the classic backfitting method and the Bayesian framework. To achieve optimal accuracy-complexity tradeoff, we extend this model with a novel variant of projection pursuit regression. Our primary result -- projection pursuit Gaussian Process Regression -- shows orders of magnitude speedup while preserving high accuracy. The natural second and third steps include non-Gaussian observations and higher dimensional equispaced grid methods.  We introduce novel techniques to address both of these necessary directions.  We thoroughly illustrate the power of these three advances on several datasets, achieving close performance to the naive Full GP at orders of magnitude less cost.
\end{abstract}

\begin{IEEEkeywords}
 Gaussian Processes, Backfitting, Projection-Pursuit Regression, Kronecker matrices.
\end{IEEEkeywords}}

\maketitle

\IEEEdisplaynotcompsoctitleabstractindextext

%
\IEEEpeerreviewmaketitle

\section{Introduction}
Gaussian Processes (GP) have become a popular tool for nonparametric Bayesian regression. Naive GP regression has $\mathcal{O}(N^{3})$ runtime (matrix inversions and determinants) and $\mathcal{O}(N^2)$ memory complexity, where $N$ is the number of observations. At ten thousand or more, this problem is for all practical purposes intractable, given current hardware.

A significant amount of research has gone into sparse approximations (reducing run-time complexity to ${\mathcal{O}(M^{2}N)}$ for some ${M \ll N}$). For an excellent review of sparse GP approximations, see \cite{Quinonero-Candela2005}.
All sparse approximation methods are based on the assumption of conditional independence of the training and test sets, given an active set of inducing inputs. As emphasized in \cite{Quinonero-Candela2005}, the results of these algorithms can depend strongly on the properties of the data. Since different assumptions fit different datasets, and since sparsity has by no means solved all efficiency issues for GPs, it is imperative to explore alternative avenues for attaining scalability.

The central aim of this paper is to introduce \emph{structured} GPs for multidimensional inputs. Specifically we present three novel advances which allow efficient and sometimes exact inference, or at least a superior runtime-accuracy tradeoff than existing methods. We say a GP is \emph{structured} if its marginals $p(\mbf{f}|\mbf{X}, \theta)$ contain exploitable structure that enables reduction in computational complexity. While these structured GP methods are known in the case of scalar inputs, many regression applications involve multivariate inputs. Our main contribution is three nontrivial extensions of these algorithms to deal with this case.

\subsection{Gaussian Process Regression}
In brief, GP regression is a Bayesian method for nonparametric regression, where a prior distribution over continuous functions is specified via a Gaussian process. (the use of GP in machine learning is well described in \cite{Rasmussen2006}).

A GP is a distribution on $f$ over an input space $X$ such that any finite selection of input locations $\mbf{x}_{1},\dots,\mbf{x}_{N}\in X$ gives rise to a multivariate Gaussian density over the associated targets, i.e.,
\begin{equation}
p(f(\mbf{x}_1),\dots,f(\mbf{x}_N)) = \mathcal{N}(\mbf{m}_N, \mbf{K}_N),
\end{equation}
where $\mbf{m}_N = m(x_1,\ldots,x_N)$ is the mean vector and $\mbf{K}_N = \{k(x_i,x_j)\}_{i,j}$ is the covariance matrix for mean function $m$ and covariance function $k$. In this paper we are specifically interested in the basic equations for GP regression, which involve two steps.  First, for given data $\mbf{y}$ (making the standard assumption of zero-mean data, without loss of generality), we calculate the predictive mean and covariance at $M$ unseen inputs as:
\begin{eqnarray}
\label{eq:gprmudef}
\bm{\mu}_{\star} & = & \mbf{K}_{MN}\left(\mbf{K}_{N} + \sigma_{n}^{2}\bm{I}_{N}\right)^{-1}\mbf{y},  \\
\label{eq:gprSigmadef}
\bm{\Sigma}_{\star} & = & \mbf{K}_{M} - \mbf{K}_{MN}\left(\mbf{K}_{N} + \sigma_{n}^{2}\bm{I}_{N}\right)^{-1}\mbf{K}_{NM},
\end{eqnarray}
For model selection, since the function $k(\cdot,\cdot; \theta)$ is parameterized by hyperparameters such as amplitude and lengthscale (which we group into $\theta$), we must consider the log marginal likelihood $Z(\theta)$:
\begin{eqnarray}
\log Z(\theta)  &=&  -\frac{1}{2}\left(\mbf{y}^{\top}(\bm{K}_{N} + \sigma_{n}^{2}\bm{I}_{N})^{-1}\mbf{y} \right.\nonumber		\\
&+& \left.\log(\det((\bm{K}_{N} + \sigma_{n}^{2}\bm{I}_{N}))) + N\log(2\pi)\right).
\end{eqnarray}
Here we use this marginal likelihood to optimize over the hyperparameters in the usual way \cite{Rasmussen2006}. The runtime of GP regression and hyperparameter learning is $\mathcal{O}(N^3)$ due to $\left(\mbf{K}_{N} + \sigma_{n}^{2}\bm{I}_{N}\right)^{-1}$, which is present in all equations.


\subsection{Gauss-Markov Processes}\label{sec:gmp}
We briefly review the use of Gauss-Markov Processes for efficient GP regression on scalar inputs, as a starting point for the multidimensional extensions in Section\ \ref{sec:multi}. Although the Gauss-Markov Process are well studied, their use for exact and efficient GP regression is under-appreciated.   A GP with a kernel corresponding to a state-space model can be viewed as a Gauss-Markov Process, enabling linear runtime. Gauss-Markov Processes can be viewed as the solution of an order-$m$ linear, stationary stochastic differential equation (SDE), given by:
\begin{equation}
\frac{d^{m}f(x)}{dx^{m}} + a_{m-1}\frac{d^{m-1}f(x)}{dx^{m-1}} + \dots + a_{1}\frac{df(x)}{dx} + a_{0}f(x) = w(x), \label{eq:sde}
\end{equation}
where $w(x)$ is a zero-mean white noise process. Note that $x$ can be any scalar input, including time. Because $w(x)$ is a GP and $f$ is linear in its coefficients, $f$ is also a GP. See \cite{Arnold1992} for an excellent introduction to SDEs. We can rewrite Eq.\ (\ref{eq:sde}) as a {vector Markov process}:
\begin{equation}
\frac{d\mbf{z}(x)}{dx} = \mbf{A}\mbf{z}(x) + \mbf{L}w(x),\label{eq:vectormp}
\end{equation}
where
\begin{eqnarray}
\mbf{z}(x) &=& \left[f(x), \frac{df(x)}{dx}, \dots, \frac{d^{m-1}f(x)}{dx^{m-1}}\right]^{\top},
\end{eqnarray}
and where $\mbf{L} = \left[0, 0, \dots, 1\right]$, and $\mathbf{A}$ is the usual suitable coefficient matrix. Eq.\ (\ref{eq:vectormp}) shows that, given knowledge of $f(x)$ and its first $m$ derivatives, we have Markov structure in the graph underlying GP inference, which will enable all efficiency gains in this section.

Earlier work \cite{Hartikainen2010,Saatci2011}, derived the SDEs corresponding to several commonly used covariance functions including the Mat\'{e}rn family and spline kernels, and good approximate SDEs corresponding to the exponentiated-quadratic kernel. Once the SDE is known, the Kalman filtering \cite{Kalman1960} and Rauch-Tung-Striebel (RTS) \cite{Rauch1965} smoothing algorithms (which correspond to belief propagation) can be used to perform GP regression in $\mathcal{O}(N)$ time and memory, a noteworthy leap in efficiency. Note that the Gauss-Markov Process framework requires sorted input points. Thus, if a sorting step is required to preprocess the data, the runtime complexity will be $\mathcal{O}(N \log N)$.  However, as this is not always relevant and certainly not the focus of the algorithms presented here, we assume the inputs are sorted in advance and refer to these models as $\mathcal{O}(N)$ in runtime complexity. For this paper we will summarize the usual system equations as:

\begin{IEEEeqnarray}{rCl}
&&\textbf{Initial state : }  p(\mbf{z}(x_{1}))  = \mathcal{N}(\mbf{z}(x_{1}); \mbf{\mu}_1, \mbf{V}_1), \label{eq:initst} \\
&&\textbf{State update : }  p(\mbf{z}(x_{i}) | \mbf{z}(x_{i-1}))  \nonumber \\
&&\hspace{2cm}= \mathcal{N}(\mbf{z}(x_{i}); \mbf{\Phi}_{i-1}\mbf{z}(x_{i-1}), \mbf{Q}_{i-1}),\\
&&\textbf{Emission : }  p(y(x_{i}) | \mbf{z}(x_{i})) \label{eq:emission}  \nonumber \\
&&\hspace{2cm}= \mathcal{N}(y(x_{i}); \mbf{h}^T\mbf{z}(x_{i}), \sigma_{n}^{2}),
\end{IEEEeqnarray}
where we assume that the inputs $x_i$ are sorted in ascending order, and the system matrices $\mbf{\Phi}$ and $\mbf{Q}$ are functions of the original GP's hyperparameters. Using these update and emission equations in the standard Kalman or RTS framework allows exact regression over the Gauss-Markov Process for a single-input dimension.

\section{Structured GP on Multiple Input Dimensions}\label{sec:multi}
Despite its importance for a variety of applications, tractable extension of the state-space models (Section\ \ref{sec:gmp}) to higher-dimensional input spaces has not been addressed in the literature. Doing so involves a number of novel steps and is the primary contribution of this work. We introduce three new algorithms for structured GPs over multivariate inputs, namely: (Sec.\ \ref{sec:multiSSM}) additive multidimensional regression (with extension to additive covariates), (Sec.\ \ref{sec:gamgp}) non-Gaussian likelihood extensions, and (Sec.\ \ref{sec:kron}) GPs over a multidimensional grid. Full details of the development of these algorithmic extensions, including important proofs, can be found in our supporting work \cite{Saatci2011}.

\subsection{GP Regression for Multidimensional State-Space Models}
\label{sec:multiSSM}
For the purposes of extending one-dimensional Gauss-Markov Processes (Sec.\ \ref{sec:gmp}) to multiple dimensions, we use the assumption of additivity.  The optimal accuracy-efficiency tradeoff will be presented in Section\ \ref{sec:ppgp}. Here we introduce the building blocks. The resulting model regresses a sum of $D$ Gauss-Markov Processes (which are independent {a priori}), where $D > 1$ is the dimensionality of the input space. Additive GP regression can be described using the following generative model:
\vspace{-0.1in}

\bieee{rCl}
y_i & = & \sum_{d = 1}^{D} \bm{f}_{d}(X_{i,d}) + \epsilon \qquad i = 1,\dots,N,  \label{eq:addgp1}\\
\bm{f}_d(\cdot) & \sim & \mathcal{GP}\left(\bm{0}; k_{d}(\bm{x}_d, \bm{x}_d';\theta_d)\right) \qquad d = 1,\dots,D, \label{eq:addgp2} \\
\epsilon & \sim & \mathcal{N}(0, \sigma_{n}^{2}), \label{eq:addgp3}
\eieee
where $X_{i,d}$ is the $d$-th component of input $i$, $k_{d}(\cdot, \cdot)$ is the kernel of the scalar GP along dimension $d$, $\theta_d$ represent the dimension-specific hyperparameters, and $\sigma_{n}^{2}$ is the (global) noise hyperparameter. Although interactions between input dimensions are not modeled a priori, an additive model does offer {interpretable} results -- one can simply plot the posterior mean of the individual $\bm{f}_d$ to visualize how each predictor relates to the target \cite{Duvenaud2011}.

We introduce a novel multidimensional Gauss-Markov Process regression where the underlying algorithm can be viewed as a Bayesian interpretation of the classical {backfitting} method \cite{Breiman1985,Hastie1990}. As described in \cite{Hastie2009}, a nonparametric regression technique (such as the spline smoother) which allows a scalable fit over a scalar input space can be used to fit an additive model over a $D$-dimensional space with the same overall asymptotic complexity, by means of the backfitting algorithm.

Surprisingly, the application of backfitting (Algorithm\ \ref{alg:bfgp}) can be proved to converge to the exact posterior mean. The easiest way to see this is by viewing (Algorithm\ \ref{alg:bfgp}) as a Gauss-Seidel iteration. We detail this in-depth proof in our supporting work \cite{Saatci2011}. As a reminder, Gauss-Seidel is an iterative technique to solve linear systems, in this case solving for the exact posterior mean (see \cite{Bertsekas1989} for a great Gauss-Seidel reference). It is precisely the additive Gauss-Markov Process structure that makes the backfitting update equivalent to a Gaussi-Seidel step.

\begin{algorithm}
	\caption{Efficient Computation of Additive GP Posterior Mean via Backfitting}\label{alg:bfgp}
	\SetKwComment{Comment}{}{}
	\SetKwInOut{Input}{inputs}
	\SetKwInOut{Output}{outputs}
	\Input{Training data $\left\{\bm{X}, \mbf{y}\right\}$. Suitable covariance function. Hypers $\theta = \bigcup_{d = 1}^{D} \{\theta_d\} \cup \sigma_{n}^{2}$.}
	\Output{Posterior training means: $\sum_{d=1}^{D} \bm{\mu}_{d}$, where $\bm{\mu}_{d} \equiv \mathbb{E}(\bm{f}_d|\bm{y}, \bm{X}, \theta_{d}, \sigma_{n}^{2})$.}
	\BlankLine
	Zero-mean the targets $\bm{y}$\;
	Initialise the $\bm{\mu}_d$ (e.g. to $\bm{0}$)\;
	\While{The change in $\bm{\mu}_d$ is above a threshold} {
		\For{$d = 1,\dots,D$} {
			$\bm{\mu}_d \la \mathbb{E}(\bm{f}_{d} | \bm{y} - \sum_{j \neq d} \bm{\mu}_{j}, \bm{X}_{:,d}, \theta_d, \sigma_{n}^{2})$\Comment*{$\triangleright$ Use state-space model here.}
		}
	}
\end{algorithm}

\begin{figure}[htb]
\center{\includegraphics[scale=0.65]{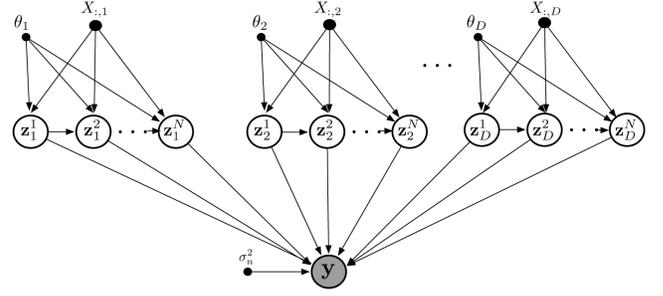}}
\caption[Graphical Model of Additive Regression using a sum of state-space models.]{\small{Graphical model for efficient additive GP regression. Each dimension is written in its corresponding state-space model.}}
\label{fig:additive-vb}
\end{figure}

To calculate posterior variances and learn hyperparameters, we must investigate further. We can express the underlying graphical model as in Fig.\ \ref{fig:additive-vb}, where we have made the state-space representation of each scalar Gauss Markov process explicit. The observed variables are the targets $\bm{y}$, and the latent variables $\bm{Z}$ consist of the $D$ Markov chains:
\beq
\bm{Z} \equiv \left(\underbrace{\bm{z}_1^{1}, \dots, \bm{z}_{1}^{N}}_{\equiv \bm{Z}_1}, \underbrace{\bm{z}_2^{1}, \dots, \bm{z}_{2}^{N}}_{\equiv \bm{Z}_2}, \dots, \underbrace{\bm{z}_D^{1}, \dots, \bm{z}_{D}^{N}}_{\equiv \bm{Z}_{D}}\right).\label{eq:zdef}
\eeq

Unfortunately, the true posterior $p(\bm{Z}_1, \dots, \bm{Z}_D | \bm{y}, \bm{X}, \theta)$ is hard to handle computationally because all variables $\bm{Z}_i$ are coupled in the posterior. Although everything is still Gaussian, we are no longer able to use the efficient state-space methods of Section\ \ref{sec:gmp} returning us to the original computational intractability at large $N$. Thus, we require an approximate inference technique such as variational Bayesian expectation maximization (VBEM) or Markov Chain Monte Carlo (MCMC) \cite{Bishop2007}. We now briefly introduce our use of these well-known technologies, as the details will demonstrate the important connection to the backfitting algorithm. Note, that the main benefits of using these algorithms comes from their scalability as they are able to inherit the linear time complexity of the state-space model.

\subsubsection{Variational-Bayesian Expectation Maximization} \label{sec:gpvb}
{\bf{E-Step}}: We use a variational-Bayesian (VB) approximation to the E-step by making the standard assumption of an approximate posterior that factorizes across the $\bm{Z}_i$, i.e.:
   \beq
   q(\bm{Z}) = \prod_{i=1}^{D} q(\bm{Z}_i).  \label{eq:vbgp}
   \eeq
Given such a factorized approximation, it can be shown that $\text{KL}(q(\bm{Z}) || p(\bm{Z} | \bm{y}, \theta))$ can be minimized in an iterative fashion, using the following central update rule \cite{Bishop2007}:
\beq
\log q(\bm{Z}_j) = \mathbb{E}_{i \neq j}(\log p(\bm{y}, \bm{Z}|\theta)) + \text{const. }\label{eq:vb-update}
\eeq
where $\mathbb{E}_{i \neq j}(\cdot)$ is an expectation with respect to $\prod_{i \neq j} q(\bm{Z}_i)$. Using Eqs.\ (\ref{eq:vbgp}) and (\ref{eq:vb-update}), we derive the iterative updates required for VBEM. We first write down the log joint over all variables, given by:
\bieee{rCl}
\label{eq:logjoint}
\log(p(\bm{y}, \bm{Z} | \theta)) = &\sum_{n=1}^{N}& \log p\left(y_n | \bm{h}^T\sum_{d=1}^{D}\bm{z}_{d}^{t_{d}(n)}, \sigma_{n}^{2}\right) + \nonumber \\ &\sum_{d=1}^{D}&\sum_{t=1}^{N}\log p(\bm{z}_{d}^{t}|\bm{z}_{d}^{t-1}, \theta_d),
\eieee
where we have defined $p(\bm{z}_{d}^{t}|\bm{z}_{d}^{t-1}, \theta_d) \equiv p(\bm{z}_{d}^{1} | \theta_d)$, for $t=1$, and $\bm{h}^T\bm{z}$ gives the first element of $\bm{z}$. Note that it is also necessary to define the mapping $t_d(\cdot)$ which gives, for each dimension $d$, the state-space model index associated with $y_n$. The index $t$ iterates over the {sorted} input locations along axis $d$. Because the expectation of the right hand side of Eq.\ (\ref{eq:logjoint}) does not depend on $\bm{z}_j$, we can consider that the mean of the Gaussian in the first term of Eq.\ (\ref{eq:logjoint}), allowing us to write:
\vspace{-0.1in}

\bieee{rCl}
&&\log q(\bm{Z}_j)\nonumber \\
&& \hspace{0.5cm} =  \sum_{n=1}^{N} \log \mathcal{N}\left(\left(y_{n} - \bm{h}^T\sum_{i \neq j} \mathbb{E}\left[\bm{z}_{i}^{t_{d}(n)}\right]\right); \bm{h}^T\bm{z}_{j}^{t_{j}(n)}, \sigma_{n}^{2}\right)\nonumber\\
&& \hspace{0.5cm} +  \sum_{t=1}^{N} \log p(\bm{z}_{j}^{t}|\bm{z}_{j}^{t-1}, \theta_j) + \text{const. } \label{eq:state-space modelbf}
\eieee
where
\beq
\mathbb{E}\left[\bm{z}_{i}^{k}\right] = \int \bm{z}_{i}^{k}q(\bm{Z}_i) d\bm{Z}_i.
\eeq
A key and somewhat surprising outcome of Eq.\ (\ref{eq:state-space modelbf}) is that in order to update the factor $q(\bm{Z}_j)$ in the E step, it is sufficient to run the standard state-space model inference procedure using only the pseudo-observations: $\left(y_{n} - \bm{h}^T\sum_{i \neq j} \mathbb{E}\left[\bm{z}_{i}^{t_{d}(n)}\right]\right)$.

A number of conclusions can be drawn from this connection. First, since VB iterations are guaranteed to converge, any moment computed using the factors $q(\bm{Z}_i)$ is also guaranteed to converge. Convergence of these moments is important because they are used to learn the hyperparameters. Second, since the {true} posterior $p(\bm{Z}_1, \dots, \bm{Z}_D | \bm{y}, \theta)$ is a large joint Gaussian over all the latent variables, $\mathbb{E}_{q}(\bm{Z})$ will be {equal} to the true posterior mean. This is because the true posterior is Gaussian (unimodal with the mean as its mode) and the VB approximation is mode-seeking \cite{Minka2005}. Additionally, as is typical for variational methods, the posterior covariance will be underestimated because $\text{KL}(q(\bm{Z}) || p(\bm{Z} | \bm{y}, \theta))$  is an \emph{exclusive} divergence measure \cite{Minka2005}.

The central VB update is precisely a backfitting update, thus illustrating a novel connection between approximate Bayesian inference for additive models and classical estimation techniques. Furthermore, this provides an alternative proof of why backfitting computes exact posterior means over latent function values.

{\bf{M-Step}}: We must optimize $\mathbb{E}_{q}\left(\log p(\bm{y}, \bm{Z}|\theta)\right)$ over $\theta$. Using Eq.\ (\ref{eq:logjoint}) it is easy to show that the expected sufficient statistics required to compute derivatives with respect to $\theta$ are the set of expected sufficient statistics for the state-space model associated with each individual dimension. This separability is another major advantage of using the factorized approximation to the posterior. Thus, for every dimension $d$, we use the Kalman filter and RTS smoother to compute $\left\{\mathbb{E}_{q(\bm{Z}_d)}(\bm{z}_{d}^{n})\right\}_{n=1}^{N}$, $\left\{\mathbb{V}_{q(\bm{Z}_d)}(\bm{z}_{d}^{n})\right\}_{n=1}^{N}$ and $\left\{\mathbb{E}_{q(\bm{Z}_d)}(\bm{z}_{d}^{n}\bm{z}_{d}^{n+1})\right\}_{n=1}^{N-1}$. We then use these expected statistics to compute derivatives of the expected complete data log-likelihood with respect to $\theta$ and use a standard minimizer (we use a conjugate gradient method) to complete the M step. 

\subsubsection{Markov Chain Monte Carlo (MCMC)} \label{sec:MCMC}
An important and customary comparison to VB is MCMC, which carries the usual benefits of approximate hyperparameter integration, but at a reduced efficiency \cite{Neal1993}. Here we briefly discuss our fairly standard MCMC implementation, noting only the important differences.


As in standard MCMC, we extended the model to include a prior over the hyperparameters. The hyperparameters for each univariate function $\mbf{f}_d$ are given a prior parameterized by $\{\mu_l, v_l, \alpha_\tau, \beta_\tau\}$, where $\{\mu_l, v_l\}$ correspond to the covariance function hyperparameter $\ell$ and $\{\alpha_\tau, \beta_\tau\}$ to $\tau_d$. We also place a $\Gamma(\alpha_n, \beta_n)$ prior over the noise precision hyperparameter $\tau_n$. We run Gibbs sampling where we block-sample the latent chains. The algorithm used to sample from the latent Markov chain in a state-space model has been called the forward-filtering, backward sampling algorithm, where forward filtering is followed by a backward sampling from the conditionals $p(\mbf{z}_k|\mbf{z}^{sample}_{k+1}; \mbf{y}; \mbf{X}_{:,d}; \theta_d)$ \cite{Douc2009}. The sampling is initialized by sampling from $p(z_K|\mbf{y};\mbf{X}_{:,d}; \theta_d)$, which is computed in the final step of the forward filtering run, to produce $z^{sample}_K$. The forward-filtering, backward sampling algorithm generates a sample of the entire state vector jointly (over training and test input locations).

%

\subsubsection{Efficient Projected Additive GP Regression}\label{sec:ppgp}
So far, we have shown how the assumption of additivity can be exploited to derive non-sparse GP regression algorithms which scale as $\mathcal{O}(N)$. These considerable efficiency gains can however decrease accuracy and predictive power versus a full unstructured GP, due to the limited expressivity of the simple additive model. To address this, we now demonstrate a relaxation of the additivity assumption \emph{without} sacrificing the $\mathcal{O}(N)$ scaling, by considering an additive GP regression model in a feature space linearly related to original space of covariates \cite{Vivarelli1998,Snelson2006}. We call this algorithm projection pursuit Gaussian Process regression (PPGPR).

\begin{figure}[htb]
\center{\includegraphics[scale=0.90]{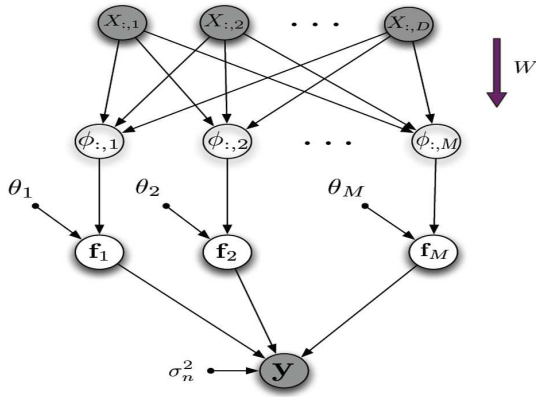}}
\caption[Graphical model for Projected Additive GP Regression.]{\small{Graphical model for Projected Additive GP Regression. In general, $M \neq D$. We present a greedy algorithm to select $M$, and jointly optimize $W$ and $\{\theta_{m}\}_{m=1}^{M}$.}}
\label{fig:ppgpr}
\end{figure}

We show that learning and inference for such a model can be performed by using \emph{projection pursuit} GP regression, a novel fusion of the classical projection pursuit regression algorithm with GP regression, with no change to computational complexity. The graphical model illustrating this idea is given in Figure\ \ref{fig:ppgpr}. We refer to the following \emph{projected additive} GP prior:
\vspace{-0.1in}
\bieee{rCll}
y_i & = & \sum_{m = 1}^{M} \bm{f}_{m}(\bm{\phi}_{m}(i)) + \epsilon & i = 1,\dots,N, \label{eq:ppadd} \\
\bm{\phi}_{m} & = & \bm{X}\bm{w}_{m}, & \\
\bm{f}_m(\cdot) & \sim & \mathcal{GP}\left(\bm{0}; k_{m}(\bm{\phi}_m, \bm{\phi}_m';\theta_m)\right) \quad & m = 1,\dots,M, \label{eq:ppr} \\
\epsilon & \sim & \mathcal{N}(0, \sigma_{n}^{2}). & \nonumber
\eieee
Notice that the number of projections, $M$, can be less or greater than $D$. Forming linear combinations of the inputs before feeding them into an additive GP model significantly enriches the flexibility of the functions supported by the prior above, including many terms which are formed by taking {products} of covariates, and thus can capture relationships where the covariates jointly affect the target variable. In fact, Eqs.\ (\ref{eq:ppadd}) through to (\ref{eq:ppr}) are identical to the standard neural network model where the nonlinear activation functions are modeled using GPs.

For inference and learning, we now derive a novel greedy algorithm which is similar to another classical nonparametric regression technique known as \emph{projection pursuit} regression \cite{Friedman1981}.

Consider the case where $M = 1$. In this case, the resulting projected additive GP regression model reduces to a scalar GP with inputs given by $\bm{X}\bm{w}$. Recall from Section\ \ref{sec:gmp} that, for a kernel that can be represented as a state-space model, we can use the EM algorithm to optimize $\theta$ with respect to the marginal likelihood efficiently, for some fixed $\bm{w}$. It is possible to extend this idea and jointly optimize $\bm{w}$ and $\theta$ with respect to the marginal likelihood, although we opt to optimize the marginal likelihood directly. Notice that every step of this optimization scales as $\mathcal{O}(N)$, since at every step we need to compute the marginal likelihood of a scalar GP (and its derivatives). These quantities are computed using the Kalman filter by differentiating the Kalman filtering process with respect to $\bm{w}$ and $\theta$. All that is required is the derivatives of the state transition and process covariance matrices ($\bm{\Phi}_{t}$ and $\bm{Q}_{t}$, for $t = 1,\dots,N-1$) with respect to $\bm{w}$ and $\theta$.

We now handle the case where ${M>1}$ using a greedy approach. At each iteration we find the optimal projection weights $\bm{w}$. The greedy nature of the algorithm allows the learning of the dimensionality of the feature space, $M$, rather naturally -- one keeps on adding new feature dimensions until there is no significant change in performance (e.g., normalized mean-squared error). One important issue which arises involves the initialization of  $\mbf{w}_{m}$ at step $m$. In our simulations we chose to initialize the weights as those obtained from a linear regression of $\bm{X}$ onto the target/residual vector $\bm{y}^{m}$. This method acts as an educated guess expecting a faster convergence rate. We call this algorithm, which learns $W$ and $\theta$, projection pursuit GP regression (PPGPR).  To be clear, note that PPGPR is used for the purposes of learning $W$ and $\theta$ only. Once this is complete, one can simply run the VB E-step to compute predictions. The PPGPR algorithm offers a bridge between the flexibility and elegance of the naive full GP, and the efficiency of its approximate additive counterpart. Its strength lies not only in the connection to backfitting, but also in its substantial improvement in performance and efficiency, as will be shown in Section\ \ref{sec:results}.

\subsubsection{Parallelization of State-Space Models}
The above algorithms can be parallelized to achieve even more speed up. By transforming the full GP to a state-space model formulation, we replaced calculating an inverse of a large joint covariance matrix to that of manipulating much smaller evolution and emission matrices ($\mbf{\Phi}$, $\mbf{Q}$), for all input locations. As these matrices are functions of the hyperparameters, they must be recalculated in every iteration during the hyperparameters learning stage. Notice, however, that for a fix set of hyperparameters, the values of  $\mbf{\Phi}$, and $\mbf{Q}$ for all locations are independent, and hence can be precalculated in parallel. We used a very simple parallelization scheme across (up to) 8 worker threads. We will further discuss the gains this in Section\ \ref{sec:multiReg}. It is important to note that as the speed of the CPUs has come to a halt and the number of cores is on the rise, the ability to use parallel schemes will be a must for any efficient GP algorithm in the future.

\subsection{Generalized Additive GP Regression}\label{sec:gamgp}
So far we have shown new methods for scaling multidimensional GP regression. Here, we extend the efficiency of these methods to include problems that need non-Gaussian likelihood functions, such as classification. Being able to efficiently handle large multidimensional datasets of non-Gaussian distributed targets is especially important in classification problems, which regularly have multiple input features, and a discrete label. Inference and learning with non-Gaussian emissions necessitates the use of approximations to the true posterior, e.g., expectation propagation (EP) \cite{Minka2001}, and Laplace approximation (LA) \cite{Rasmussen2006}. However, only a handful of works in the literature focus on tractable algorithms for big data where exact GP classification is intractable. These works are extensions of the sparse GP framework using various approximation methods \cite{Guzman2007,Vanhatalo2007,Lawrence2003}. Sparse GP classification, however, is not without problems: for example, pseudo input learning with EP and hyperparameter learning expand the parameter space making the problem more susceptible to overfitting. Other works which use a subset of data are more stable; however, this can affect accuracy as the active subset may not be optimal for the sparse conditional independence assumption. Additionally, there is the problem of model selection for the number of points to use in the sparse active set. Thus, extending our structured GP model to the non-Gaussian case is necessary and useful.

Here we derive, for additive structured GP kernels, an $\mathcal{O}(N)$ algorithm which performs MAP inference and hyperparameter learning using Laplace's approximation. The resulting LA algorithm is both stable and accurate. Note, that although it has been suggested in literature that the EP approximation has superior approximation qualities (for binary classification \cite{Nickisch2008}), the choice of LA is due to its computational benefits as, to our knowledge, there is no good method for lowering the computational complexity of the EP updates from $\mathcal{O}(N^3)$. We now derive this LA algorithm, and then we end this section with another key insight showing that this algorithm is a Bayesian interpretation of another classical technique known as local scoring \cite{Hastie1986}.

Given the likelihood $p(\bm{y}|\bm{f})$ is non-Gaussian, we use the standard Laplace approximation:
\beq
p(\bm{f} | \bm{y}, X, \theta) \approxeq \mathcal{N}(\hat{\bm{f}}, \Lambda^{-1}),
\eeq
where ${\hat{\bm{f}} \equiv \arg\max_{\bm{f}}p(\bm{f} | \bm{y}, X, \theta)}$ and the approximated covariance matrix ${\Lambda \equiv -\nabla\nabla \log p(\bm{f} | \bm{y}, X, \theta) |_{\bm{f} = \hat{\bm{f}}}}$. We define the following objective:
\beq
\Omega(\bm{f}) \equiv \log p(\bm{y} | \bm{f}) + \log p(\bm{f} | X, \theta). \label{eq:laobj}
\eeq
$\hat{\bm{f}}$ is found by applying Newton's method to this objective. Newton's method is guaranteed to converge to the global optimum, given the objective in Eq.\ (\ref{eq:laobj}) is convex with respect to $\bm{f}$. This is the case for many cases of generalized regression, as the likelihood term is usually log-concave, as well as the prior term.

If we assume that $\bm{f}$ is drawn from an additive GP, then it follows that the required gradient and Hessian (for Newton iterations) are:
\vspace{-0.1in}

\bieee{rCl}
\nabla\Omega(\bm{f}) & = & \nabla_{\bm{f}} \log p(\bm{y} | \bm{f}) - \bm{K}_{\text{add}}^{-1}\bm{f}, \\
\nabla\nabla\Omega(\bm{f}) & = & \underbrace{\nabla\nabla_{\bm{f}} \log p(\bm{y} | \bm{f})}_{\equiv -W} - \bm{K}_{\text{add}}^{-1}.
\eieee
This makes the Newton iteration:
\bieee{rCl}
\bm{f}^{(k+1)} & \la & \bm{f}^{(k)} + \left(\bm{K}_{\text{add}}^{-1} + W\right)^{-1} \nonumber\\
&&\cdot \left(\nabla_{\bm{f}} \log p(\bm{y} | \bm{f})|_{\bm{f}^{(k)}} - \bm{K}_{\text{add}}^{-1}\bm{f}^{(k)}\right), \\
& = & \bm{K}_{\text{add}}\left(\bm{K}_{\text{add}} + W^{-1}\right)^{-1} \nonumber \\
&&\cdot \left[\bm{f}^{(k)} + W^{-1}\nabla_{\bm{f}}\log p(\bm{y} | \bm{f})|_{\bm{f}^{(k)}}\right].\label{eq:cool}
\eieee
Since the generalized additive GP model assumes conditional independence of $\bm{y}$ given $\bm{f}$, $p(\bm{y} | \bm{f}) = \prod_{n=1}^{N} p(y_i | f_i)$, and $\bm{W}$ is a diagonal matrix and therefore easy to invert. Looking closer at Eq.\ (\ref{eq:cool}), we see that it is \emph{precisely} the same as the expression to compute the posterior mean of a GP, where the target vector is given by $\left[\bm{f}^{(k)} + \bm{W}^{-1}\nabla_{\bm{f}}\log p(\bm{y} | \bm{f})|_{\bm{f}^{(k)}}\right]$ and where the diagonal \qq{noise} term is given by $\bm{W}^{-1}$. Given an additive kernel corresponding to a sum of scalar GPs that can be represented using state-space models, we can therefore use Algorithm\ \ref{alg:bfgp} to implement a single iteration of Newton's method. As a result, it is possible to compute $\hat{\bm{f}}$ in $\mathcal{O}(N)$ time, since in practice only a handful of Newton iterations are required for convergence. Wrapping backfitting iterations inside a global Newton iteration is precisely how the local-scoring algorithm is run to fit a generalized additive model \cite{Hastie1986}. Thus, we can view the development in this section as a novel Bayesian interpretation of local scoring.

The above calculates the posterior Laplace approximation.  To efficiently approximate the marginal likelihood, we use the Taylor expansion of the objective function $\Omega(\bm{F})$, although we will need to express it explicitly in terms of $\bm{F} \equiv [\bm{f}_1; \dots; \bm{f}_{D}]$, as opposed to the sum over $\bm{f}$.
\beq
\Omega(\bm{F}) = \log p(\bm{y} | \bm{F}) + \log p(\bm{F} | \bm{X}, \theta).
\eeq
Once $\hat{\bm{F}}$ is known, it can be used to compute the approximation to the marginal likelihood. Using first-order Taylor expansion we obtain:
\vspace{-0.1in}

\bieee{lCl}
&& \hspace{-0.2 cm}\log p(\bm{y} | \bm{X}) \nonumber \\
&& \hspace{0.2 cm} \approx  \Omega(\hat{\bm{F}}) - \frac{1}{2}\log\det\left(\tilde{\bm{W}} + \tilde{\bm{K}}^{-1}\right) + \frac{ND}{2} \log(2\pi) \label{eq:taylor2}\\
&&\hspace{0.2 cm} =  \log p(\bm{y} | \hat{\bm{F}}) - \frac{1}{2}\hat{\bm{F}}^{\top}\tilde{\bm{K}}^{-1}\hat{\bm{F}} \nonumber \\
&&\hspace{0.2 cm} -  \frac{1}{2}\log\det\left(\tilde{\bm{K}} + \tilde{\bm{W}}^{-1}\right) - \frac{1}{2}\log\det(\tilde{\bm{W}}), \label{eq:mdl}
\eieee
where $\tilde{\bm{K}}$ is a block diagonal tiling of $\bm{K}_1,\bm{K}_2,\ldots,\bm{K}_D$, and $\tilde{\bm{W}}$ is a block diagonal tiling of the single $\bm{W}$ matrix. We used the matrix determinant lemma to get from Eq.\ (\ref{eq:taylor2}) to Eq.\ (\ref{eq:mdl}). Importantly, all the the terms in Eq.\ (\ref{eq:mdl}) can be computed in $\mathcal{O}(N)$ runtime due to the fact that the latent function can be represented as a sum of state-space models (for more details see \cite{Saatci2011}) .

In summary, we showed that by using the Laplace approximation, we are able to maintain low runtime complexity by combining a Newton method with additive regression update in Eq.\ (\ref{eq:cool}) (local scoring), and by approximating the marginal likelihood using Eq.\ (\ref{eq:taylor2}). 

\subsection{Gaussian Processes on Multidimensional Grids}\label{sec:kron}
The second GP kernel structure that can be exploited involves the assumption of {equispaced inputs}. This is commonly seen for GP regression in time and space (e.g., regular measurements at evenly spaced weather stations, or video captured by a CCD camera). Even though there are good Toeplitz methods for scalar equispaced inputs \cite{Cunningham2008}, the extension to a multidimensional grid has not been addressed in literature. In this section, we present a novel method to perform exact inference in $\mathcal{O}(N)$ time for any tensor product kernel (most commonly-used kernels are of this form), using properties of Kronecker products.

In this case, we can compute all the computationally troublesome quantities involved (such as $(\bm{K}_{N} + \sigma_{n}^{2}\bm{I}_{N})^{-1}\bm{y}$) using a few matrix-vector products (of size $N$). Importantly, these matrix-vector products have the form:
\beq
\mbf{\alpha} = \left(\bigotimes_{d=1}^{D} \mbf{A}_{d}\right)\mbf{b},\label{eq:mvprod}
\eeq
all of which can be computed in linear runtime.

Computing $\mbf{\alpha}$ using standard matrix-vector multiplication is an $\mathcal{O}(N^2)$ operation. However, with problems of this form, it is possible to attain linear runtime using tensor algebra \cite{Riley2006}. For the relevant background in tensor algebra, see the appendix in our supporting work \cite{Saatci2011}. The product in Eq.\ (\ref{eq:mvprod}) can be viewed as a tensor product between the tensor $\bm{T}^{A}_{i_{1},j_{1},\dots,i_{D},j_{D}}$ representing the {outer product} over $[\bm{A}_{1},\dots,\bm{A}_{D}]$, and $\bm{T}^{B}_{j_{D},\dots,j_{1}}$. The term $\bm{T}^{B}_{j_{D},\dots,j_{1}}$ represents the length-$N$ vector $\mbf{b}$. Conceptually, the aim is to compute a {contraction} over the indices $j_{1}, \dots, j_{D}$, namely:
\beq
\bm{T}^{\alpha} = \sum_{j_{1}=1}^{G_1}\dots\sum_{j_{D}=1}^{G_1} \bm{T}^{A}_{i_{1},j_{1},\dots,i_{D},j_{D}}\bm{T}^{B}_{j_{D},\dots,j_{1}}, \label{eq:tensorprod}
\eeq
where $\bm{T}^{\alpha}$ is the tensor representing the solution vector $\bm{\alpha}$. As the sum in Eq.\ (\ref{eq:tensorprod}) is over ${N=\prod_{d=1}^D G_d}$ elements, it will run in $\mathcal{O}(N)$ time.
Equivalently, we can express this operation as a sequence of matrix-tensor products and tensor {transpose} operations.
\begin{equation}
\mbf{\alpha} = \operatorname{vec}\left(\left(\mbf{A}_{1}\dots\left(\mbf{A}_{D-1}\left(\mbf{A}_{D}\bm{T}^{B}\right)^{\top}\right)^{\top}\right)^{\top}\right),
\label{eq:tensorp}
\end{equation}
where we define matrix-tensor products of the form $\mbf{Z} = \mbf{X}\mbf{T}$ as:
\begin{equation}
\mbf{Z}_{i_{1}\dots i_{D}} = \sum_{k}^{\operatorname{size(\mbf{T},1)}} \mbf{X}_{i_{1}k}\mbf{T}_{ki_{2}\dots i_{D}}.
\end{equation}
The operator $\top$ is assumed to perform a cyclic permutation of the indices of a tensor, namely
\begin{equation}
\mbf{Y}^{\top}_{i_{D}i_{1}\dots i_{D-1}} = \mbf{Y}_{i_{1}\dots i_{D}}.
\end{equation}
Furthermore, when implementing the expression in Eq.\ (\ref{eq:tensorp}) it is possible to represent the tensors involved using matrices where the first dimension is retained and all other dimensions are collapsed into the second, resulting in a matrix $\mbf{B}$ which is a $G_{d}$-by-$\prod_{j \neq d}G_{j}$ matrix, where $G_d$ is the number of elements in dimension $d$.
Algorithm\ \ref{alg:kron-mvprod} gives pseudo-code illustrating these steps. In short, we use tensor algebra to create a linear-runtime method (Algorithm\ \ref{alg:kron-mvprod}) for doing matrix-vector multiplications across a Kronecker product matrix, which arises quite naturally for most GP kernels on a grid of inputs.

\begin{algorithm}
\caption{Efficient matrix-vector multiply for Kronecker matrices}\label{alg:kron-mvprod}
\SetKwFunction{Reshape}{reshape}
\SetKwFunction{Vectorize}{vec}
\SetKwFunction{Size}{size}
\SetKwComment{Comment}{}{}
\SetKwInOut{Input}{inputs}
\SetKwInOut{Output}{outputs}
\Input{$D$ matrices $[\mbf{A}_{1}\dots\mbf{A}_{D}]$, length-$N$ vector $\mbf{b}$}
\Output{$\mbf{\alpha}$, where $\mbf{\alpha} = \left(\bigotimes_{d=1}^{D} \mbf{A}_{d}\right)\mbf{b}$}
\BlankLine
$\mbf{x} \leftarrow \mbf{b}$\;
\For{$d \leftarrow D$ \KwTo $1$}{
	$G_d \leftarrow $ \Size{$\mbf{A}_{d}$}\;
	$\mbf{X} \leftarrow $ \Reshape{$\mbf{x}$, $G_d$, $N/G_{d}$}\;
	$\mbf{Z} \leftarrow \mbf{A}_{d}\mbf{X}$ \hspace{0.5cm} \Comment{$\triangleright$ Matrix-tensor product}
	$\mbf{Z} \leftarrow \mbf{Z}^{\top}$  \hspace{0.8cm}  \Comment{$\triangleright$ Tensor rotation}
	$\mbf{x} \leftarrow $ \Vectorize{$\mbf{Z}$}\;
}
$\mbf{\alpha} \leftarrow \mbf{x}$\;
\end{algorithm}

The critical second step is to note that $(\mbf{K} + \sigma^2_n \mbf{I}_N)^{-1}$ cannot itself be written as Kronecker product, due to the perturbation on the main diagonal. Nevertheless, it is possible to
sidestep this problem using the eigendecomposition properties and the identity
\begin{equation}\label{eq:kroneig}
\left(\mbf{K} + \sigma_{n}^{2}I_{N}\right)^{-1}\mbf{y} = \mbf{Q}\left(\mbf{\Lambda} + \sigma_{n}^{2}I_{N}\right)^{-1}\mbf{Q}^{\top}\mbf{y}.
\end{equation}
Importantly, the eigenvector matrix $\mbf{Q}$ will also be a Kronecker product.  Hence, to then efficiently solve Eq.\ (\ref{eq:kroneig}), we first evaluate and perform eigendecompositions of covariances along {individual} dimensions to get $[\mbf{Q}_{d}, \mbf{\Lambda}_{d}]$. This has complexity $\mathcal{O}((\max_{d} G_{d})^{3})$, which is negligible compared to $N = \prod_{d=1}^{D} G_{d}$. Next, we calculate Eq.\ (\ref{eq:kroneig}) in three steps:

\bieee{rCl}
\mbf{\alpha} &\leftarrow& \Kron\left([\mbf{Q}_{1}^{\top},\dots,\mbf{Q}_{D}^{\top}], \mbf{y}\right), \\
\mbf{\alpha} &\leftarrow& \left(\mbf{\Lambda} + \sigma_{n}^{2}I_{N}\right)^{-1}\mbf{\alpha}, \\
\mbf{\alpha} &\leftarrow&  \Kron\left([\mbf{Q}_{1},\dots,\mbf{Q}_{D}], \mbf{\alpha}\right),
\eieee
where we efficiently used $\texttt{kron\_mvprod}$ (Alg.\ \ref{alg:kron-mvprod}) twice and noting that the matrix $\mbf{\Lambda} + \sigma_{n}^{2}I_{N}$ is easy to invert as it is diagonal. Computation of the test set predictions $\mbf{Q}\left(\mbf{\Lambda} + \sigma_{n}^{2}I_{N}\right)^{-1}\mbf{Q}^{\top}\mbf{K}_{MN}$, can be done efficiently using the same approach. The result is the third main contribution of this paper: a linear-runtime\footnote{The complexity of the algorithm is ${\mathcal{O}\left( \left(\log_D N\right)^3 N \right)}$, however it rapidly converges to $\mathcal{O}(N)$ as the number of dimensions grows.} method for calculating the key regression equation $\left(\mbf{K} + \sigma_{n}^{2}I_{N}\right)^{-1}\mbf{y}$ using only the assumption that the inputs lie on a grid.

In summary, we exploited two important realizations: efficient eigendecomposition using properties of the Kronecker product, and tensor products enabling fast multiplication by matrices that can be written as a Kronecker product. This novel improvement for exact GP inference opens the door to a whole new set of applications, which would have never been considered otherwise, such as GP on images or videos.

\section{Results}\label{sec:results}
Here we will compare the algorithms discussed in the paper to other commonly used algorithms both in the GP world, and other common machine learning techniques.

\subsection{Multidimensional Regression} \label{sec:multiReg}
In this section we will compare methods for multidimensional regression on both simulated and real experimental data. For each experiment presented, we will compare both runtime and accuracy. If a particular algorithm has a stochastic component to it (e.g., if it involves MCMC) its performance will be averaged over 10 runs. Every experiment was composed of training (i.e., smoothing and hyperparameter learning given $\{\bm{X}, \bm{y}\}$) and testing phases. In each experiment, we used 1000 points for test sets.

In terms of accuracy, we use two standard performance measures: normalized mean square error (NMSE) and test-set Mean Negative Log Probability (MNLP).
\bieee{rCl}
\text{NMSE} &=& \frac{\sum_{i=1}^{N_{\star}} (\bm{y}_{\star}(i) - \bm{\mu}_{\star}(i))^{2}}{\sum_{i=1}^{N_{\star}} (\bm{y}_{\star}(i) - \bar{y})^{2}}, \nonumber\\
\text{MNLP} &=& \frac{1}{2N_{\star}}\sum_{i=1}^{N_{\star}}\left[\frac{(\bm{y}_{\star}(i) - \bm{\mu}_{\star}(i))^{2}}{\bm{v}_{\star}(i)} + \log \bm{v}_{\star}(i) + \log 2\pi\right], \nonumber
\eieee
where $\bm{\mu}_{\star} \equiv \mathbb{E}(\bm{f}_{\star} | X, \bm{y}, X_{\star}, \theta)$, $\bm{v}_{\star}\equiv \mathbb{V}(\bm{f}_{\star} | X, \bm{y}, X_{\star}, \theta)$, and $\bar{y}$ is the training-set average target value. These measures have been chosen to be consistent with those commonly used in the sparse GP regression literature.
We compare runtime performance in seconds, taking into account both the learning and prediction phases.

We test the following algorithms (with the following names): the full naive GP implementation (Full GP), additive models (Sections\ \ref{sec:gpvb} and \ref{sec:MCMC}) using VBEM inference (Additive-VB) and the MCMC inference (Additive-MCMC), projected additive models using greedy projection pursuit of Section\ \ref{sec:ppgp} (PPGPR-Greedy) and a variation of MCMC (PPGPR-MCMC). Finally, for the sparse GP method we used the sparse pseudo-input Gaussian process (SPGP) \cite{Snelson2006}. However, to be conservative, we did not learn the pseudo inputs (which can potentially greatly increase the algorithm complexity and runtime) but rather used a random subset of the inputs as the active set. For both the SPGP and the Full-GP, we used the GPML Matlab Code version 3.1 \cite{Rasmussen2010}. Also note that, for {Additive-VB} and {PPGPR-greedy} we have set the number of outer loop iterations (the number of VBEM iterations for the former, and the number of projections for the latter) to be at maximum 10 for all $N$. Increasing this number increased the cost with no change to accuracy, so this is a reasonable choice. All algorithms were run both as a single thread and using a parallel multicore, but since SPGP and and Full-GP do not offer efficient implementation of the parallel schemes, their results were the same for both cases\footnote{When discussing parallel schemes we refer to only the learning stage. As in all GP frameworks, parallelism can always be used for prediction, since we are only interested in the predictive marginals per test point. However, this does not have any noticeable effect on the runtime and is thus unimportant to the comparison.}.

\subsubsection{Synthetic Data Experiments} \label{sec:regsynth}
First we used synthetic data generated by the following model:
\bieee{rcll}
y_i & = & \sum_{d = 1}^{D} \bm{f}_{d}(x_{:,d}) + \epsilon & i = 1,\dots,N, \label{eq:ppradd} \\
\bm{f}_d(\cdot) & \sim & \mathcal{GP}\left(\bm{0}; k_{d}(\bm{x}_d, \bm{x}_d';[1, 1])\right) \quad & d = 1,\dots,D, \label{eq:ffbsme} \\
\epsilon & \sim & \mathcal{N}(0, 0.01), & \nonumber
\eieee
where $k_{d}(\bm{x}_D, \bm{x}_d';[1, 1])$ is given by the Mat\'{e}rn(7/2) kernel with unit lengthscale and amplitude. We used $D=8$ dimensions, and collected runtimes for a set of values for $N$ ranging from 1000 to 50000.

\begin{figure}[htb]
\centering
{\includegraphics[width=0.5\textwidth]{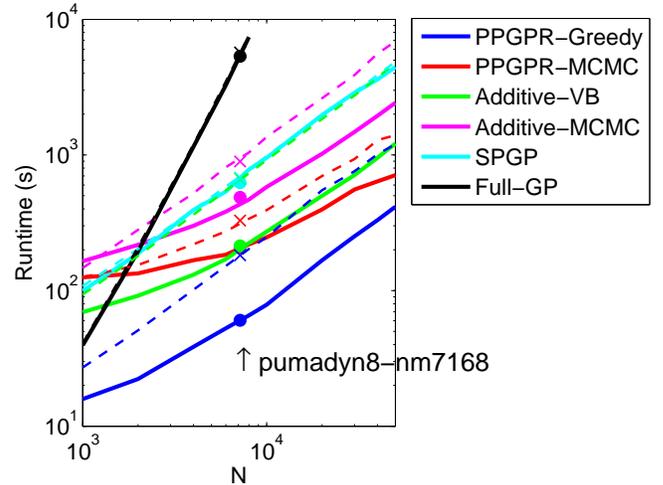}}
\caption{A comparison of runtimes for efficient Bayesian additive GP regression, with $D=8$, $N = [2; 4; 6; 8; 10; 20; 30; 40; 50]\times10^3$, presented as a log-log plot. The algorithms ran on a Linux server, once as a single thread (dash lines) and once in a multicore parallel scheme using 8 processors (solid lines). At N=7168, we added an overlay of the runtime results for the pumadyn8-nm dataset (described in Section\ \ref{sec:regreal}) for both single ('x') and multicore ('o') runs.}
\label{reg_runtime_N_comp_8D}
\end{figure}

Figure\ \ref{reg_runtime_N_comp_8D} illustrates the significant computational savings attained by exploiting the structure of the additive kernel. To find the relationship between the number of inputs to the runtime, we calculated a linear slope of the data in the log-log scale. As expected, the slope of the Full-GP is close to three due to its cubic complexity, and all the approximation algorithms have runtimes that scale linearly with the input size. We can also see that parallel processing of the state-space model matrices offers further improvement in scaling. These results serve only as a rough estimate, because the performance can depend on the chosen algorithm parameters, such as: number of outer loop iterations in the Additive-VB, number of projections in PPGPR-greedy, or number of samples in the MCMC methods. This runtime/accuracy consideration should be used when comparing the efficiency of the algorithms.

Additionally, runtime on a modern computer is by no means a perfect measure of algorithmic complexity. Nonetheless, we will see that the results of Fig.\ \ref{reg_runtime_N_comp_8D} agree with all the results from the real datasets. For example, in Fig.\ \ref{reg_runtime_N_comp_8D} we overlay the results of one of the real datasets, and one sees a close correspondence between synthetic and real data. Thus, these and subsequent results are highly representative and assert the primary point of this section: the runtime of our approximation algorithms do indeed scale linearly with $N$, versus the cubic scaling of the naive GP implementation, not GP-grid.

\begin{figure}[htb]
\vspace{-10pt}
\center{\includegraphics[width = 0.5\textwidth]{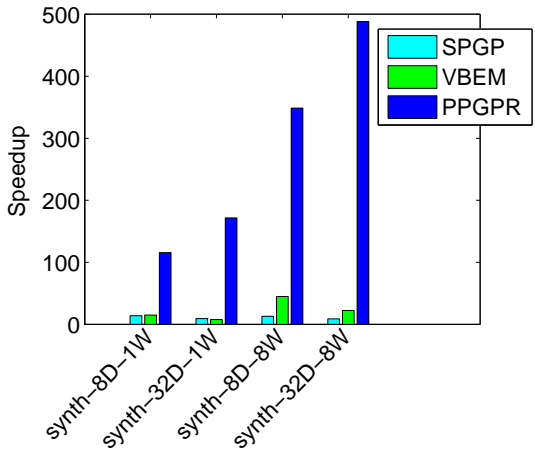}}
\caption{A comparison of the speed up offered by the approximation algorithms compared with exact GP. The runtime was measured on the learning stage for three approximation algorithms: sparse GP, Additive VB, and greedy Projection-Pursuit. The comparison was done using synthetic results with different dimensions (8D and 32D), and running on both a single and multicore (8-core) computer. As can be seen from the figure, the greedy Projection-Pursuit offers the highest speedup, and is especially efficient in high dimensions.}
\label{fig:speedup}
\end{figure}

Fig.\ \ref{fig:speedup} shows the effects of increased dimensionality on the approximate algorithms. In this figure we show the runtime speedup of the algorithms with respect to the runtime of the Full-GP on the synthetic data generated with dimensionality of either $D=8$ or $D=32$. In all the runs the number of inputs was set to $N=8000$, and the algorithms were run once with a single thread (1 worker = 1W), and once using the parallel scheme (8 workers = 8W). In the multidimensional case, the projection pursuit algorithm exhibits the largest speedup, as it allows for a reduction in the number of effective dimensions (via the greedy selection). Notably, PPGPR-Greedy achieves consistently an order of magnitude improvement over SPGP.

\subsubsection{Real Data Experiments} \label{sec:regreal}
\begin{figure}[htb]
\vspace{-10pt}
\centering
\subfloat[]{\label{fig:reg_compare_full}\includegraphics[width = 0.50\textwidth]{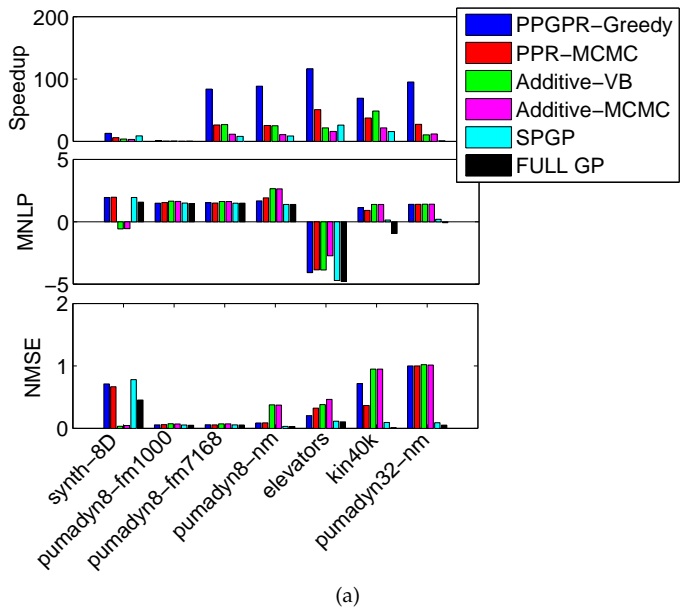}}
\caption{These figures offer a comparison between the different GP methods discussed in the text, taking into account both speedup and accuracy. For comparison we used several known datasets from literature and ran the algorithms on a multicore (8-core) computer. The top figure illustrates the speedup of the approximation algorithms runtimes with respect to the full GP (exact inference) runtime. The bottom two figures show two metrics for calculating regression accuracy.}
\label{fig:reg_compare}
\end{figure}

Next, we extend the comparison to real datasets, which will allow thorough accuracy comparisons. We test over seven well-known datasets.  These data sets are: {\tt synth-8D} ($N = 8000$ synthetic data from Section\ \ref{sec:regsynth}). Next, the {\tt pumadyn} family is a robotic arm dataset from \cite{Rasmussen2006}, and consist of three datasets: {\tt pumadyn8-fm1000} ($N=1000$, fairly linear data with $D=8$ dimensions), {\tt pumadyn8-fm7168} ($N=7168$, fairly linear data with $D=8$ dimensions), {\tt pumadyn32-nm} ($N=7168$, highly nonlinear data with $D=32$). {\tt Elevators} dataset consists of the current state of the f16 aircraft ($N = 8752$, 17-dimensional) \cite{Lazaro-Gredilla2010}, and {\tt kin40k} is a highly nonlinear dataset ($N=10000$, 8-dimensional) as introduced in \cite{Seeger2003}. Fig.\ \ref{fig:reg_compare} demonstrates the central analysis of this section. In each subplot, we calculate speedup, MNLP, and NMSE across all seven datasets and six algorithmic options. To reiterate, the two additive and two PPGPR algorithms are our advances of Section\ \ref{sec:multiSSM}, and our main comparison points are SPGP \cite{Snelson2006} and a naive full GP implementation.  The top subplot in Fig.\ \ref{fig:reg_compare} indicates the substantial speedups offered by all algorithms over the full GP, with the exception only of the $N=1000$ dataset ({\tt pumadyn8-fm1000}; this is not surprising given small N). Further, as indicated in Figure\ \ref{reg_runtime_N_comp_8D}, our PPGPR-Greedy achieves the largest speedup across all datasets, and in most cases the error (MNLP and NMSE) is the same as competing methods. The first four or five datasets tell a very similar accuracy story across PPGPR-Greedy, SPGP, and the full GP. We also see that the simple additive models almost always underperform in accuracy, which is as expected given their limited expressivity compared to PPGPR-Greedy. The one exception where Additive-VB outperforms PPGPR-Greedy is the synthetic data set. However, this is expected as we used an additive model to generate data and the greedy nature of PPGPR-Greedy causes it to underperform. In the final two datasets, we see that SPGP and the full GP have considerably better accuracy. This may be explained as both these datasets are highly nonlinear, making the additive assumption inaccurate. 

\begin{figure}[htb]
\vspace{-20pt}
\center{\includegraphics[width = 0.5\textwidth]{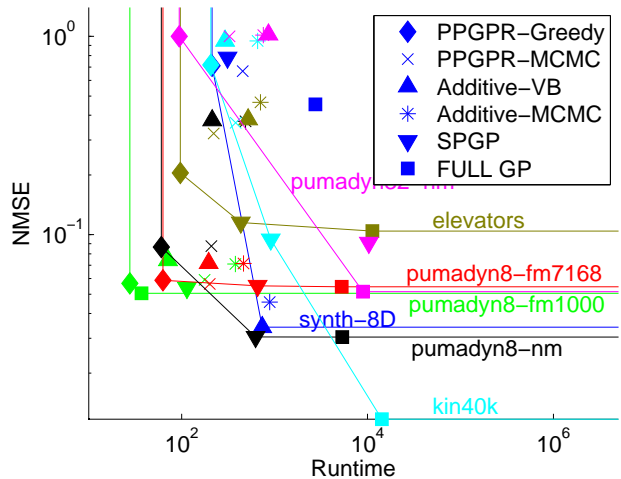}}
\caption{The two fundamental desiderata of our algorithms are accuracy and speed. Here we plot error vs runtime to quantify the tradeoff between these two objectives using the notion of Pareto efficiency. Every algorithm is represented using a unique marker and with a color scheme chosen according to the datasets. For each dataset, the Pareto efficient frontier is shown as a color line passing through the efficient algorithms for that dataset.}
\label{fig:efficientFront}
\end{figure}

Understanding the runtime-accuracy tradeoff based on problem requirements is essential. As we just described, PPGPR-greedy achieves the best runtimes but at times with an accuracy cost. Thus we want to quantify the notion of a runtime-accuracy tradeoff. To do so we plot all data sets and algorithms in a runtime vs. error plot (Fig.\ \ref{fig:efficientFront}), and we use the economics concept of Pareto efficiency: \emph{efficient} points in the runtime vs error plot represent algorithms with minimum runtime for a given error rate. Pareto inefficient algorithms are then those points that are unambiguously inferior. The efficient frontier is the convex hull of all \{runtime,error\} points (algorithms) for a given dataset.  This will give us a clear picture of which algorithms are optimal choices across a range of datasets. Fig.\ \ref{fig:efficientFront} details this, with one efficient frontier for each dataset (a given color).  Each algorithm has a given marker type. This immediately shows what one would expect: the full GP implementation is typically most accurate, but only if one is willing to invest substantial runtime. This choice is often Pareto efficient. Secondly, most often the PPGPR-greedy is the other efficient choice for a substantially reduced runtime, albeit higher error.  Surprising to note is the relative weakness of SPGP over several datasets.

In Table\ \ref{tbl:paretoFrontier} we count of the number of datasets where each algorithm is on the efficient frontier, which gives an idea of how often an algorithm is competitive with others, or optimal given a particular runtime or error budget. Three algorithms stand out in their overall efficiency: PPGPR-Greedy, SPGP, and full GP. The PPGPR-Greedy is the \emph{only} consistent efficient algorithm for \emph{all} datasets as it achieves the fastest runtime. However, more interestingly, it also achieves very good accuracy results making most other algorithms inefficient. Of course, any trivial algorithm could achieve efficiency by having minimal runtime and arbitrary error, but the data demonstrates that this is not the case with our algorithms: the PPGPR-greedy error in almost all datasets is competitive or better than all alternatives. Thus the frequent efficiency of PPGPR-greedy is legitimate.

\begin{Table}
\vskip 0.15in
\captionof{table}{Efficiency comparison, showing the number of datasets where the algorithm was on the Pareto efficient frontier. There were seven datasets tested.}
\begin{tabular}{l | c }
Algorithm & Pareto Efficient Frontier Count  \\ \hline \hline
PPGPR-Greedy & 7\\
PPGPR-MCMC & 0\\
Additive-VB & 1 \\
Additive-MCMC & 0\\
SPGP & 4 \\
Full-GP & 6 \\
\hline \hline
\end{tabular}
\vskip -0.1in
\label{tbl:paretoFrontier}
\end{Table}

\subsection{Multidimensional Classification}
In this section we will compare the generalized additive-GP from Section\ \ref{sec:gamgp} to other kernel classifiers (both Bayesian and non-Bayesian). We use common performance metrics from the sparse GP classification literature, enabling straightforward comparison with other experimental results. In this paper we will focus on the task of binary classification, however in principle, extensions to tasks such as multi-class classification and Poisson regression can be performed without affecting asymptotic complexity. For performance measures we use: algorithm runtime (in seconds), error rate, and the test-set mean negative log-likelihood (MNLL):

\bieee{rCl}
&&\text{Error Rate} = \frac{\#(\text{incorrect classifications})}{\#(\text{test cases})},\\
&&\text{MNLL} = \frac{1}{N_{\star}}\sum_{i=1}^{N_{\star}}\left[y_i \log \hat{p}_i + (1 - y_i) \log(1 - \hat{p}_i)\right].
\eieee
For both the test error rate and MNLL measures lower values indicate better performance.

\subsubsection{Synthetic Data Experiments}
We used synthetic data generated by the following model:
\bieee{rCl}
    y_i & \sim & \text{Bernoulli}(p_i)  \qquad i = 1,\dots,N, \nonumber \\
    p_i & = & g(f_i), \\
    \bm{f}(\cdot) & = & \sum_{d} \bm{f}_{d}(\cdot), \\
    \bm{f}_d(\cdot) & \sim & \mathcal{GP}\left(\bm{0}; k_{d}(\bm{x}_d, \bm{x}_d';\theta_d)\right) \qquad d = 1,\dots,D, \nonumber
\eieee
where $g(f_i)$ is the logistic link function.

\begin{figure}[htb]

\center{\includegraphics[width = 0.5\textwidth]{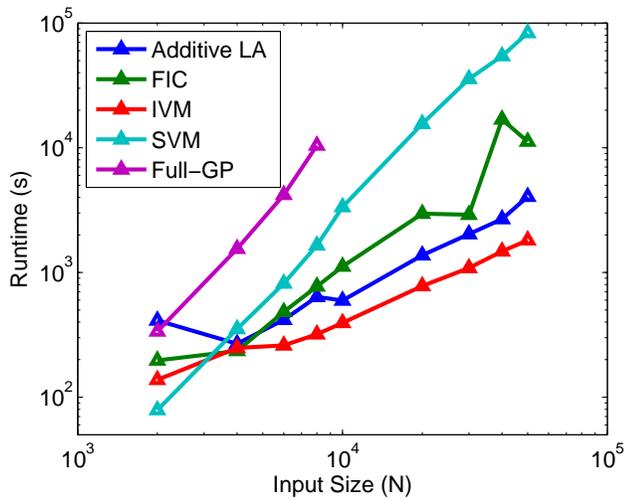}}
\caption{This figure shows the runtime of the classification algorithms for the synthetic dataset with $D=8$, $N = [2; 4; 6; 8; 10; 20; 30; 40; 50]\times10^3$. For the learning stage we used 50 iterations, and we did prediction on 1000 points. The log-log slopes of the algorithms are: Full-GP $= 2.75$, Additive-GP $= 1.07$, FIC $= 1.53$, IVM $=  0.80$, SVM $= 2.16$.}
\label{fig:class}
\end{figure}

Within the GP framework, we compared generalized additive GP Regression from Section\ \ref{sec:gamgp} (Additive-LA), standard GP classification with Laplace's approximation (Full-GP) \cite{Rasmussen2006}, and sparse GP methods of informative vector machine (IVM) \cite{Lawrence2003} and fully independent conditional (FIC) \cite{Quinonero-Candela2005}. For completeness, we also include support vector machines (SVM) \cite{Cortes1995}.\footnote{To calculate the MNLL, we used the probabilistic predictions from the SVM using cross-validation and cross-entropy metric \cite{Platt1999}} For the Full-GP we used GPML Matlab Code version 3.1 \cite{Rasmussen2010}; for FIC we used the GPstuff Matlab package \cite{Vanhatalo2012}; for SVM we used LIBSVM \cite{LIBSVM}; and for IVM we used the implementation given in \cite{Lawrence2003}. We tested the algorithms on the synthetic data from the model above using 8 dimensions while varying the number of inputs $N = [2; 4; 6; 8; 10; 20; 30; 40; 50]\times10^3$. We stopped running the Full-GP at $10000$ as it took too long to finish. A comparison of the runtime results is shown in Fig.\ \ref{fig:class}. To be consistent, we used exactly 25 iterations for all algorithms during the learning stage. As can be seen from the figure, Additive-GP offers excellent scaling for large input sizes. The only algorithm that offers faster runtime than the additive-GP is IVM. This can be expected as the IVM only uses the information in the active set and discards the rest. Our algorithm, on the other hand, makes use of all the data, and is thus able to achieve a more accurate estimation, as the results below demonstrate.

\subsubsection{Real Data Experiments}
We tested the classification algorithms from the previous section on three additional popular datasets: {\tt Breast Cancer} \cite{Mangasarian1990}, {\tt Magic Gamma Telescope} \cite{Bock2004}, and {\tt IJCNN} \cite{ijcnn2001}. We again only allowed 25 iterations in the learning stage. For sparse methods we tested two activeset sizes: 50, and $0.1N$. Table\ \ref{tbl:class-results} summarizes the classification results across all algorithms and datasets. Each column gives the classification error rate, MNLL, and runtime. First, we consider the runtime. We note that, as expected from Figure\ \ref{fig:class}, the Full GP has the least attractive runtime in all but the {\tt Breast Cancer} data (due to the small data size). The IVM has the best runtime performance across all datasets, after which our Additive-LA method is superior. In terms of performance, error rates are fairly consistent throughout the data sets, with the exception of notably high errors for IVM in the last two data sets. This is echoed by MNLL, where the IVM tends to have a significantly larger error. These results are not as compelling as in the regression case (as is often the case when comparing Bayesian methods to an SVM), but the Additive-LA is competitive overall. Thus, if a Bayesian method is needed for nonparametric classification, the Additive-LA approach is a viable and stable solution.

\begin{Table}
\captionof{table}{Performance Comparison of efficient Bayesian additive GP classification algorithms with commonly-used classification techniques on larger datasets.}
\vskip 0.15in
\begin{center}
\begin{tabular}{l | c | c | c}
Algorithm & Error Rate & MNLL & Runtime (s) \\ \hline \hline
\multicolumn{4}{c}{\textit{Synthetic Additive Data ($N =  4000$, $M =  1000$, $D =     8$)}} \\ \hline
Full-GP & 0.6040 & 0.7402 & 2244.5111 \\
{Additive-LA} & $\quad 0.2800 \quad$ & $\quad 0.5929 \quad$ & 161.1185 \\
FIC -    50 & 0.2800 & 1.0640 &525.6684 \\
FIC -   400 & 0.4550 & 1.3612 &850.8427 \\
IVM -    50 & 0.2800 & 0.6931 & 65.1951 \\
\hline
SVM & 0.2940 & 0.5838 & 345.7267 \\
\hline \hline
\multicolumn{4}{c}{\textit{Breast Cancer ($N =   359$, $M =    90$, $D =     9$)}} \\ \hline
Full-GP & 0.0667 & 0.1436 & 6.0435 \\
{Additive-LA} & $\quad 0.0667 \quad$ & $\quad 0.1215 \quad$ & 89.2993 \\
FIC -    50 & 0.0556 & 0.0999 &27.9584 \\
FIC -    36 & 0.0556 & 0.0999 &55.9182 \\
IVM -    50 & 0.0667 & 0.6382 & 12.4867 \\
\hline
SVM & 0.0556 & 3.1717 & 1.7062 \\
\hline \hline
\multicolumn{4}{c}{\textit{Magic Gamma Telescope ($N = 15216$, $M =  3804$, $D =    10$)}} \\ \hline
Full-GP & NA & NA & NA \\
{Additive-LA} & $\quad 0.1393 \quad$ & $\quad 0.3419 \quad$ & 2345.6546 \\
FIC -    50 & 0.1441 & 0.3656 &3339.6185 \\
FIC -  1522 & 0.1396 & 0.3654 &7331.2780 \\
IVM -    50 & 0.6583 & 0.6932 & 118.0407 \\
\hline
SVM & 0.1191 & 0.3026 & 8070.2400 \\
\hline \hline
\multicolumn{4}{c}{\textit{IJCNN ($N = 49990$, $M = 91701$, $D =    13$)}} \\ \hline
Full-GP & NA & NA & NA \\
{Additive-LA} & $\quad 0.0516 \quad$ & $\quad 0.1560 \quad$ & 14505.5000 \\
FIC -    50 & 0.0482 & 1.1859 &4390.2498 \\
FIC -  4999 & 0.0770 & 0.8171 &16728.0911 \\
IVM -    50 & 0.0950 & 0.6932 & 369.0000 \\
\hline
SVM & 0.0166 & 0.0509 & 22170.1000 \\
\hline \hline
\end{tabular}
\end{center}
\vskip -0.1in
\label{tbl:class-results}
\end{Table}

\subsection{Multidimensional Regression on a Grid}
In this section we consider data with input on a multidimensional grid. We compare the exact GP-grid method from Section\ \ref{sec:kron} to the naive Full-GP method and show an application for this method in image reconstruction.

\begin{figure}[htb]
\center{\includegraphics[width=0.50\textwidth]{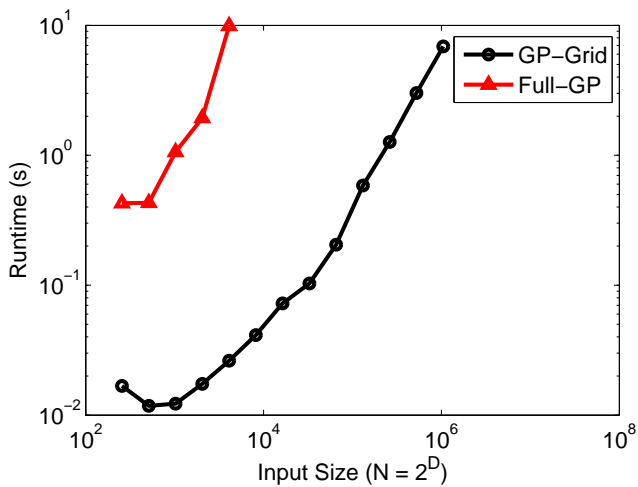}}
\caption{Runtimes of naive Full-GP in red and GP regression on Grid using Kronecker product (GP-Grid from Sec.\ \ref{sec:kron}) in black, for $N = [2^8; 2^9; \ldots ; 2^{20}]$. The slope for the naive Full-GP is 2.6 and that of Kronecker product is 1.05 (based on last 7 points). This empirically verifies the improvement to linear scaling.}
\label{fig:kron}
\end{figure}

We first start by comparing the runtime complexity of the GP-grid to Full-GP. For this we ran both algorithms on synthetic data where each synthetic dataset has input locations at the corners of the $\{-1,1\}$ hypercube in D dimensions. The target value is set to noise.  At each run we increased the dimension $D$ by 1, thereby multiplying the number of input points by 2. Fig.\ \ref{fig:kron} illustrates the time it takes for one iteration during the learning stage. As can be seen, the GP-grid scales linearly with the input size while the Full-GP is cubic.

\begin{figure}[htb]
\vspace{-10pt}
\centering
\subfloat[]{\label{fig:GPMLpicOrig}\includegraphics[width = 0.23\textwidth]{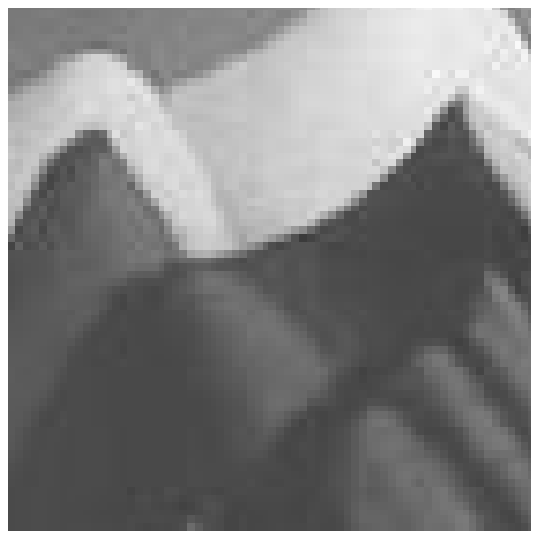}}
\subfloat[]{\label{fig:GPMLpicGPR}\includegraphics[width = 0.23\textwidth]{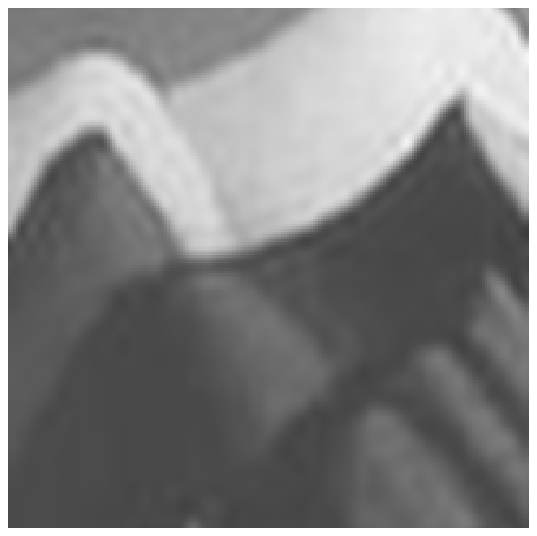}}\\
\subfloat[]{\label{fig:GPMLpicDS}\includegraphics[width = 0.23\textwidth]{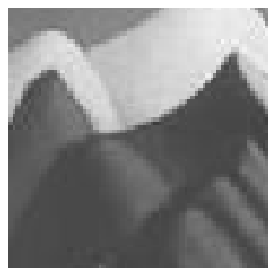}}
\subfloat[]{\label{fig:GPMLpicDS_GPR}\includegraphics[width = 0.23\textwidth]{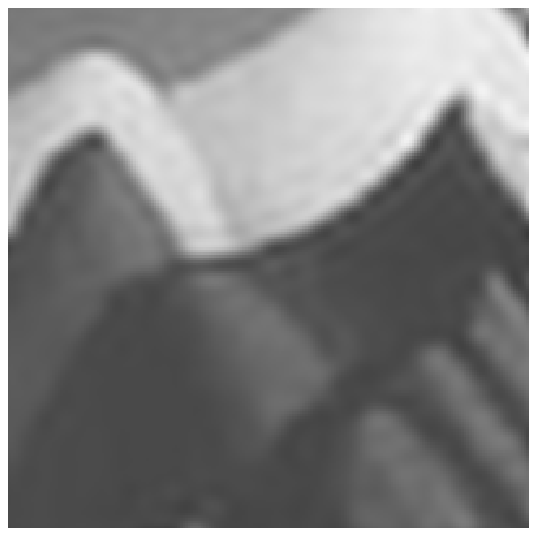}}
\caption{Illustration of GP regression on a picture as an equidistance grid. Fig.\ \ref{fig:GPMLpicOrig} shows a $200 \times 200$ pixels of the original noisy picture. in Fig.\ \ref{fig:GPMLpicGPR} we used the GPR-grid method from Section\ \ref{sec:kron} to learn the parameters and reconstruct the picture. Fig.\ \ref{fig:GPMLpicDS} is a down-sampled the original picture (Fig.\ \ref{fig:GPMLpicOrig}) by 2 in both dimensions ($\frac{1}{4}$ of the data). Fig.\ \ref{fig:GPMLpicDS_GPR} shows a reconstruction of the picture where the GPR-grid used the down-sampled data for learning, and then predicted the values at missing locations.}
\label{fig:GPMLpic}
\end{figure}

Improving the scalability of exact GP on a multidimensional grid may seem like an unusual special case, but it importantly enables new applications which were not tractable previously. One such application is picture reconstruction as pictures are an equispaced grid of pixels. Here we show how GP-grid can be used to reconstruct and interpolate a noisy image. We first apply GP-grid on a $200 \times 200$ pixel noisy image (Fig.\ \ref{fig:GPMLpicOrig}) and use the inferred GP posterior as a denoised reconstruction (Fig.\ \ref{fig:GPMLpicGPR}). Note that the GP nicely smooths out most of the compression artifacts that were present in the original image. Note also that this is a GP on $N=40000$, which is hopelessly intractable for Full-GP. However, GP-grid was able to learn the parameters in 42.75 seconds, and denoise in 1490.3 seconds. Further, because the two methods are mathematically proven to be identical, any small numerical differences will be due to relative instability of the naive GP implementation.

Next, we down-sampled the original image by $\frac{1}{4}$ (Fig.\ \ref{fig:GPMLpicDS}) and used GP-grid to interpolate the missing values (Fig.\ \ref{fig:GPMLpicDS_GPR}). Here the learning stage was 4.57 seconds and the reconstruction with interpolation was 186.61 seconds, and the overall quality of the interpolated reconstruction is still high. This image example is only a single example, but it is representative: due to the Kronecker method of Section\ \ref{sec:kron}, we have a provably exact GP method that scales linearly in the number of data points. Image and video analysis is a critical and common machine learning application, but the use of nonparametric Bayesian algorithms in this domain is infrequent. Our GP-grid algorithm importantly enables the use of GP technologies in this application area.
%
%

\section{Discussion and Conclusion}
Gaussian Processes are perhaps the most popular nonparametric Bayesian method in machine learning, but their adoption across other fields - and notably in application domains - has been limited by their burdensome scaling properties. Having fast, scalable methods for Gaussian Processes may mean the difference between a theoretically interesting approach and a method that is widely used in practice.

While important sparsification work has somewhat addressed this scalability issue, the problem is by no means closed. Our aim here has been to explore the use of structured GP models. We made nontrivial advances to existing state-space and equispaced GP methods in order to extend structured GP techniques into the multidimensional input domain. Our results (Section\ \ref{sec:results}) illustrate across a range of data and different algorithms that structured models are most often superior to the state of the art sparse methods (SPGP). Notably, we introduced projection pursuit Gaussian Process regression (PPGPR-greedy, Section\ \ref{sec:ppgp}), an $\mathcal{O}(N)$ runtime algorithm that combines the computational efficiency of additive GP models (Section\ \ref{sec:gpvb}) with the expressivity of a multidimensional coupled model. The result (Section\ \ref{sec:regreal}, notably Table\ \ref{tbl:paretoFrontier}) is an algorithm that has a superior runtime-accuracy tradeoff than several other algorithms including the sparse SPGP. While its accuracy was often slightly lower than a full GP, the linear scaling properties of PPGPR mean that it can be efficiently used across a much broader range of data sizes and applications. The primary takeaway of this work is thus: while the naive GP implementation may often produce the highest accuracy, the PPGPR algorithm that we introduced offers the best runtime-accuracy tradeoff across many datasets and is able to scale well beyond the realm of a naive GP.

Of course, in some cases the researcher will prefer the SPGP method over PPGPR-greedy. Indeed, in many senses, these two approaches are orthogonal to each other. We see this as an inherent fact in approximation techniques: various methods will be more appropriate in different settings. Our results (Section\ \ref{sec:results}) presented an in-depth investigation into this runtime-accuracy tradeoff, using both metrics on real datasets and meta-analyses of Pareto efficiency. Our well-founded and competitive alternatives for efficient GP regression and classification can thus enable the researcher to make an informed choice about a GP method for a given data size, data complexity, and available computational resource.

To the point of runtime-accuracy tradeoff, there are sometimes opportunities for great scaling advantages with no accuracy tradeoff whatsoever. We demonstrated such an example with equispaced inputs in Section\ \ref{sec:ppgp}. Though this method exploits structure differently than the main PPGPR-greedy algorithm, our novel use of tensor algebra to create an $\mathcal{O}(N)$  GP model belongs in this exposition of the computational advantages of careful structural consideration. Notably, this method also opens up an entirely new set of big-data applications, such as image and video processing, or financial engineering applications such as implied volatility surfaces. Our future work is pursuing these application domains.

As a last computational point, as growth in computational speed is increasingly driven by parallelism over raw processor speed, it will become increasingly important to use GP schemes that naturally incorporate parallel processing, to efficiently deal with the rapid growth of future datasets. Our PPGPR-greedy method stands out in this regard versus both the naive full GP and SPGP, and again the results of Section\ \ref{sec:results} reiterated this fact.

Finally, from an algorithmic perspective, another interesting byproduct of this work was a number of surprising connections to classical statistical techniques. The additive model turned out to be a Bayesian interpretation of the backfitting algorithm, importantly yielding an alternative proof of that algorithm's validity. We utilized another classical technique - projection pursuit - in the PPGPR model, which dramatically increased the expressivity of the additive model without sacrificing $\mathcal{O}(N)$ performance.

Understanding how our existing nonparametric models can scale and be used in real data, and how these models connect to other areas of statistics, will increase the utility of machine learning algorithms in general. This is perhaps most important with Gaussian Processes, which promise a wide range of useful applications.

\bibliographystyle{IEEEtran}
\bibliography{structuredGPBib}
\end{document}